\documentclass[sigconf]{acmart}

\usepackage{balance} 
\usepackage{booktabs} 
\usepackage{hyperref}
\usepackage{bm}
\usepackage{latexsym}

\usepackage{algorithm}
\usepackage[noend]{algorithmic}
\usepackage{multirow}
\usepackage{natbib}
\usepackage{etoolbox}
\newtoggle{conf} \toggletrue{conf}

\usepackage{pifont}

\newcommand{\cut}[1]{}

\copyrightyear{2024}
\acmYear{2024}
\setcopyright{acmlicensed}
\acmConference[KDD '24]{Proceedings of the 30th ACM SIGKDD Conference on Knowledge Discovery and Data Mining}{August 25--29, 2024}{Barcelona, Spain}
\acmBooktitle{Proceedings of the 30th ACM SIGKDD Conference on Knowledge Discovery and Data Mining (KDD '24), August 25--29, 2024, Barcelona, Spain}
\acmDOI{10.1145/3637528.3671467}
\acmISBN{979-8-4007-0490-1/24/08}

\begin{document}

\title{Grounding and Evaluation for Large Language Models: Practical Challenges and Lessons Learned (Survey)}

\author{Krishnaram Kenthapadi}
\affiliation{%
  \institution{Oracle Health AI}
  \city{Redwood City, CA}
  \country{USA}
  }
\email{krishnaram.kenthapadi@oracle.com}
\author{Mehrnoosh Sameki}
\affiliation{%
  \institution{Microsoft Azure AI}
  \city{Boston, MA}
  \country{USA}
  }
\email{mesameki@microsoft.com}

\author{Ankur Taly}
\affiliation{%
  \institution{Google Cloud AI}
  \city{Mountain View, CA}
  \country{USA}
  }
\email{ataly@google.com}

\renewcommand{\shortauthors}{Krishnaram Kenthapadi, Mehrnoosh Sameki, and Ankur Taly}

\begin{abstract}
With the ongoing rapid adoption of Artificial Intelligence (AI)-based systems in high-stakes domains, ensuring the trustworthiness, safety, and observability of these systems has become crucial.
It is essential to evaluate and monitor AI systems not only for accuracy and quality-related metrics but also for robustness, bias, security, interpretability, and other responsible AI dimensions.
We focus on large language models (LLMs) and other generative AI models, which present additional challenges such as hallucinations, harmful and manipulative content, and copyright infringement.
In this survey article accompanying our \href{https://sites.google.com/view/llm-evaluation-tutorial}{\underline{tutorial}}, we highlight a wide range of harms associated with generative AI systems, and survey state of the art approaches (along with open challenges) to address these harms.
\end{abstract}


\keywords{Responsible AI; Generative AI; Large Language Models; Grounding; Evaluations; Truthfulness; Safety and Alignment; Bias and Fairness, Model Robustness and Security; Privacy; Model Disgorgement and Unlearning; Copyright Infringement; Calibration and Confidence; Transparency and Causal Interventions.}

\maketitle

\section{Introduction}
Considering the increasing adoption of Artificial Intelligence (AI) technologies in our daily lives, it is crucial to develop and deploy the underlying AI models and systems in a responsible manner and ensure their trustworthiness, safety, and observability.
Our focus is on large language models (LLMs) and other generative AI models and applications.
Such models and applications need to be evaluated and monitored not only for accuracy and quality-related metrics but also for robustness against adversarial attacks, robustness under distribution shifts, bias and discrimination against underrepresented groups, security and privacy protection, interpretability, hallucinations (and other ungrounded or low-quality outputs), harmful content (such as sexual, racist, and hateful responses), jailbreaks of safety and alignment mechanisms, prompt injection attacks, misinformation and disinformation, fake, misleading, and manipulative content, copyright infringement, and other responsible AI dimensions.

In this tutorial, we first highlight key harms associated with generative AI systems, focusing on ungrounded answers (hallucinations), jailbreaks and prompt injection attacks, harmful content, and copyright infringement. We then discuss how to effectively address potential risks and challenges, following the framework of identification, measurement, mitigation (with four mitigation layers at the model, safety system, application, and positioning levels), and operationalization.
We present real-world LLM use cases, practical challenges, best practices, lessons learned from deploying solution approaches in the industry, and key open problems. Our goal is to stimulate further research on grounding and evaluating LLMs and enable researchers and practitioners to build more robust and trustworthy LLM applications.

We first present a brief tutorial outline in \S\ref{sec:overview}, followed by an elaborate discussion of different responsible AI dimensions in \S\ref{sec:holisticEvaluation}.
We devote \S\ref{sec:grounding} to the problem of grounding for LLM applications, and \S\ref{sec:observability} to the emerging area of ``LLM operations''. For each dimension (discussed in \S\ref{sec:holisticEvaluation} to \S\ref{sec:observability}), we present key business problems, technical solution approaches, and open challenges.

\subsection{Tutorial Overview}\label{sec:overview}
Our \href{https://sites.google.com/view/llm-evaluation-tutorial}{\underline{tutorial}} consists of the following parts:\footnote{\href{https://sites.google.com/view/llm-evaluation-tutorial}{https://sites.google.com/view/llm-evaluation-tutorial}}\\

\noindent\textbf{Introduction and Overview of LLM Applications.} 
We give an overview of the generative AI landscape in industry and motivate the topic of the tutorial with the following questions. What constitutes generative AI? Why is generative AI an important topic? What are key applications of generative AI that are being deployed across different industry verticals? Why is it crucial to develop and deploy generative AI models and applications in a responsible manner?\\

\noindent\textbf{Holistic Evaluation of LLMs.}
We highlight key challenges that arise when developing and deploying LLMs and other generative AI models in enterprise settings, and present an overview of solution approaches and open problems. We discuss evaluation dimensions such as truthfulness, safety and alignment, bias and fairness, robustness and security, privacy, model disgorgement and unlearning, copyright infringement, calibration and confidence, and transparency and causal interventions.\\

\noindent\textbf{Grounding for LLMs.}
We then provide a deeper discussion of grounding for LLMs, that is, ensuring that every claim in the response generated by an LLM can be attributed to a document in the user-specified knowledge base. We highlight how grounding differs from factuality in the context of LLMs, and present technical solution approaches such as retrieval augmented generation, constrained decoding, evaluation, guardrails, and revision, and corpus tuning.\\

\noindent\textbf{LLM Operations and Observability.}
We present processes and best practices for addressing grounding and evaluation related challenges in real-world LLM application settings. We discuss mechanisms for managing safety risks and vulnerabilities associated with deployed LLM and generative AI applications as well as practical approaches for monitoring the underlying models and systems with respect to quality and other responsible AI related metrics.
Using real-world LLM case studies, we highlight practical challenges, best practices, lessons learned from deploying solution approaches in the industry, and key open problems.

\section{Holistic Evaluation of LLMs}\label{sec:holisticEvaluation}
The overarching goal of evaluation is to determine whether \emph{a trained LLM is fit for deployment in an enterprise setting}.
A commonly quoted maxim is that LLMs must ensure \emph{helpful, truthful, and harmless} responses~\cite{askell2021general}.
While this seems straightforward, each of these dimensions has several nuances.
For instance, lack of truthfulness can range from subtle misrepresentations to making blatant false statements (colloquially known as ``hallucinations'')~\cite{huang2023survey}.
Similarly, harmful responses can vary from racially biased responses, to violent, hateful, and other inappropriate responses, to responses causing social harm (e.g., instruction on how to cheat in an examination without getting caught).
Further, in the context of evaluating LLMs, it is important to be aware of shortcomings that have been highlighted with human and automatic model evaluations and with commonly used datasets for natural language generation~\cite{gehrmann2023repairing}.

Besides evaluations of response quality, practitioners also have to worry about training data privacy, model stealing, copyright violations, and security risks such as jailbreaking~\cite{zou2023universal} and prompt injection~\cite{willison2022prompt}.
In some settings, one may also seek calibrated confidence scores for responses, interpretability, and robustness to adversarial prompts.

In the rest of this section, we outline several evaluation dimensions that arise in enterprise deployments.
Evaluation of LLMs is an important topic and there have been a number of dedicated frameworks \cite{liang2022holistic, gao10256836framework, nazir2024langtest} describing evaluation datasets, metrics, and benchmarks for various dimensions.
A growing collection of tools and resources have been proposed across different phases of LLM development~\cite{longpre2024responsible}.
Here, we focus on the key business concerns, leading solution approaches, and open challenges for each evaluation dimension.

\subsection{Truthfulness}\label{sec:truthfulness}
\noindent\textbf{Business problems}: How do we ensure that LLM responses are informed, relevant, and trustworthy? How do we detect and recover from hallucinations?\\

\noindent\textbf{Solution approaches}: There is extensive work on hallucinations in LLMs \cite{ji2023survey,huang2023survey}, including, the causes and sources of hallucinations \cite{mckenna2023sources}, and measures for evaluating LLMs based on their vulnerability to producing hallucinations \cite{rawte2023troubling}.
A variety of methods have been proposed to detect hallucinations, ranging from sampling based approaches \cite{manakul2023selfcheckgpt} to approaches leveraging internal states of the LLM \cite{snyder2023early}.
There is also early work on detecting and preventing hallucinations in large vision language models \cite{gunjal2023detecting} and other multimodal foundation models \cite{yin2023woodpecker}.

A number of methods have been proposed to fundamentally reduce hallucinations by tuning models.
One line of work involves training or fine-tuning LLMs on highly curated textbook-like datasets \cite{zhou2023lima, gunasekar2023textbooks}.
Another approach involves fine-tuning LLMs on preference data for factuality, i.e., response pairs ranked by factuality~\cite{tian2024finetuning}.
A fundamental hypothesis here is that LLMs have systematic markers for when they are being untruthful~\cite{kadavath2022language, tian-etal-2023-just}.
The fine-tuning process aims to train LLMs to tap into these markers and upweight factual responses.
Related to this, it has been conjectured that LLMs internalize different ``personas'' during pretraining, and by training on truthful question-answer pairs, one can upweight the ``truthful'' persona (even on unseen domains)~\cite{joshi2023personas}.
Reducing hallucination on a synthetic task has been explored as a way to reduce hallucination on real-world downstream tasks~\cite{jonesteaching}.
Finally, a recent work shows that fine-tuning LLMs on new information that was not acquired during pretraining can encourage the model to hallucinate~\cite{gekhman2024does}.
Curating fine-tuning sets to avoid this issue paves another path to reducing hallucinations.

While truthful responses are table stakes for enterprise deployments, we may want to go one step further and ensure that all responses are aligned with a specific knowledge base (e.g., a set of enterprise documents). This is known as \emph{grounding}.
This is a vast topic in itself, and therefore we dedicate \S\ref{sec:grounding} entirely to it.

Finally we emphasize that not all hallucinations are equally bad.
For instance, hallucinations in response to nonsensical prompts or prompts with false premises (see ~\cite{fresh_qa} for examples of questions whose premises are factually incorrect and hence ideally need to be rebutted) are relatively less concerning than hallucinations in response to well-meaning prompts.
Furthermore, hallucinations in high stakes verticals like healthcare and life sciences may be far more concerning than hallucinations in other verticals. \\

\noindent\textbf{Open challenges}: A key open challenge is detecting hallucinations in video, speech, and multimodal settings.
Another open challenge is getting LLMs to generate citations when they answer from parametric knowledge.
More specifically, can the LLM be made aware of \emph{document identifiers} during pre-training, similar to the work on differential search indexes~\cite{transformer_search_index}, so that it can generate the appropriate markers as citations for various claims in its response?
A broader challenge is to leverage ideas and lessons from search and information retrieval literature~\cite{metzler2021rethinking,zhu2023large} to improve relevance, trustworthiness, and truthfulness of LLM responses. For example, how can we incorporate valuable information such as document authors, document quality, authoritativeness of the domain, timestamp, and other relevant metadata during pre-training and subsequent stages of LLM development?

\subsection{Safety and Alignment}
\noindent\textbf{Business problems}: How do we prevent an LLM from generating toxic, violent, offensive, or otherwise unsafe output? How do we detect such content in cases where prevention fails to work? How do we ensure that the responses from an LLM are aligned with human intent even in settings where it is hard for human experts to verify such alignment?\\

\noindent\textbf{Solution approaches}: The problem can be addressed during different stages of the LLM lifecycle.
During data collection and curation, we can apply mechanisms to detect unsafe content and take remedial steps, such as excluding or modifying such content.
During pretraining and fine-tuning, we can incorporate constraints or penalties to discourage the learning of unsafe sequences.
In the reinforcement learning from human feedback (RLHF) stage, we can include response pairs with preference labels on which one is more appropriate, and tune the model to ``align'' its responses with the preferences~\cite{casper2023open}.
As part of prompt engineering, we can include instructions to discourage the LLM from generating undesirable outputs.
Finally, when prevention fails, we can apply toxicity classifiers to detect undesirable outputs (as well as undesirable inputs) and flag such instances for appropriate treatment by the user-facing AI applications.

Another direction in alignment research is leveraging more powerful LLMs to detect safety and alignment issues with a weaker LLM in a cost-effective and latency-sensitive fashion.
The problem can be framed as a constrained optimization problem: given cost or latency constraints, determine the subset of prompts and responses to be evaluated using a more powerful LLM (e.g., GPT-4). 
In certain settings, the task to be evaluated could be too hard for even human experts (e.g., comparing two different summaries of a very large collection of documents or judging the quality of hypotheses generated based on a large volume of medical literature), necessitating the use of powerful LLMs in a manner that aligns with human intent. 
The converse problem of leveraging less powerful LLMs to align more powerful LLMs with human intent has also been explored in alignment research. A related challenge is to ensure that AI systems with superhuman performance (which could possibly be smarter than humans) are designed to follow human intent.
While current approaches for AI alignment rely on human ability to supervise AI (using approaches such as reinforcement learning from human feedback), these approaches would not be feasible when AI systems become smarter than humans \cite{burns2023weak}.

Overall, alignment is an active area of research, with approaches ranging from data-efficient alignment \cite{jin2023data} to alternatives to RLHF \cite{dong2023steerlm} to aligning cross-modal representations \cite{panagopoulou2023x}.\\

\noindent\textbf{Open challenges}: There has a been a bunch of recent work on generating adversarial prompts to bypass existing mechanisms for mitigating toxic content generation~\cite{zou2023universal, wei2023jailbroken}.
A key open challenge is mitigating toxic content generation even under such adversarial prompts. 
Recent research has shown that LLM based guardrail models could themselves be attacked. For instance, a two-step prefix-based attack procedure – that operates by (a) constructing a universal adversarial prefix for the guardrail model, and (b) propagating this prefix to the response – has been shown to be effective across multiple threat models, including ones in which the adversary has no access to the guardrail model at all \cite{mangaokar2024prp}.
How do we develop effective LLM based guardrails that are robust to such attacks (and even better, have provable robustness/security guarantees)?
Another challenge lies in balancing reduction of undesirable outputs with preservation of the model's ability towards creative generation.
Finally, as LLMs are increasingly deployed as part of open-ended applications, an important socio-technical challenge is to investigate the opinions reflected by the LLMs, determine whether such opinions are aligned with the needs of different application settings, and design mechanisms to incorporate preferences and opinions of relevant stakeholders (including those impacted by the deployment of LLM based applications)~\cite{santurkar2023whose}.

\subsection{Bias and Fairness}
\noindent\textbf{Business problems}: How do we detect and mitigate bias in foundation models? How can we apply bias detection and mitigation throughout the foundation model lifecycle?\\

\noindent\textbf{Solution approaches}: There is extensive work on detecting and mitigating bias in NLP models \cite{caliskan2021detecting,caliskan_2017,bolukbasi_2016,garg2018word,de2019bias,romanov2019biasbios}. In addition to known categories of bias observed in predictive ML models, new types of bias arise in LLMs and other generative AI models, e.g., gender stereotypes, exclusionary norms, undesirable biases towards mentions of disability, religious stereotypes, and sexual objectification \cite{gallegos2023bias,bender2021dangers,wolfe2023contrastive,tamkin2023evaluating}. Additionally, due to the sheer size of datasets used, it is difficult to audit and update the training data or even anticipate different kinds of biases that may be present. Mitigation approaches include counterfactual data augmentation (or other types of data improvements), finetuning, incorporating fairness regularizers, in-context learning, and natural language instructions.
For a longer discussion, we direct the readers to the survey by Gallegos et al.~\cite{gallegos2023bias}.
More broadly, we can view bias measurement and mitigation as an important component of building a reliable and robust application that works well across different subgroups of interest (including but not necessarily limited to protected groups). By performing fine-grained evaluation and robustness testing across such groups, we can identify underperforming groups, improve the performance for such groups, and thereby potentially boost even the overall performance.\\

\noindent\textbf{Open challenges}: Bias and fairness mitigation is a relatively nascent space, and a key open question is identifying and designing practical, scalable processes from the large class of bias measurement and mitigation techniques proposed for LLMs.
A related challenge is ensuring that the bias mitigation approach does not cause the model to inadvertently demonstrate disparate treatment, which could be considered unlawful in a wide range of scenarios under US law \cite{lohaus2022two}.
Further, how do we audit LLMs and other generative AI models for different types of implicit or subtle biases and design mechanisms to mitigate or recover from such biases, although the models may not show explicit bias on standard benchmarks~\cite{haim2024biasauditingllms,bai2024measuring,hofmann2024dialect}? It has recently been argued that harmful biases are an inevitable consequence arising from the design of LLMs as they are currently formulated, and that the connection between bias and fundamental properties of language models needs to be probed further~\cite{resnik2024large}. How do we revisit the foundational assumptions underlying LLMs and approach the development and deployment of LLMs with the goal of preventing bias-related harms {\em by design}?

\subsection{Robustness and Security}
\noindent\textbf{Business problems}: How do we measure and improve the robustness of LLMs and other generative AI models and applications against minor prompt perturbations, natural distribution shifts, and other unseen or challenging scenarios? How do we safeguard LLMs against manipulative efforts by bad actors to (jail-)break alignment, reveal system prompts, and inject malicious instructions into prompts (also called \emph{prompt injection} attacks~\cite{willison2022prompt})?\\

\noindent\textbf{Solution approaches}:
Many techniques proposed for measuring and improving robustness in NLP models can be adopted or extended for LLMs. In particular, the following ideas and notions could be relevant for LLMs: definitions, metrics, and assumptions regarding robustness (such as label-preserving vs. semantic-preserving); connections between robustness against adversarial attacks and robustness under distribution shifts; similarities and differences in robustness approaches between vision and text domains; model-based vs. human-in-the-loop identification of robustness failures.
Mitigation approaches involve learning invariant representations, and ensuring models do not rely on spurious patterns using techniques like data augmentation, reweighting, ensembling, inductive-prior design, and causal intervention \cite{wang2022measure}. Open-source evaluation frameworks and benchmarks such as Stanford HELM \cite{liang2022holistic}, Eleuther Harness \cite{gao10256836framework}, LangTest \cite{nazir2024langtest}, and Fiddler Auditor \cite{fiddlerAuditorBlog} can be utilized for benchmarking different LLMs and evaluating robustness in application-specific settings.

LLMs have been shown to be vulnerable to adversarial perturbations in prompts \cite{zhu2023promptbench}, prompt injection attacks \cite{willison2022prompt}, data poisoning attacks \cite{wallace2021concealed}, and universal and transferable adversarial attacks on alignment \cite{zou2023universal}. Several benchmarks have been proposed for red-teaming  / testing LLMs against adversarial attacks and related issues \cite{ganguli2022red,perez2022red,zhu2023promptbench}. Metrics for quantifying LLM cybersecurity risks, tools to evaluate the frequency of insecure code suggestions, and tools to evaluate LLMs to make it harder to generate malicious code or aid in carrying out cyberattacks have also been proposed \cite{bhatt2023purple}. Additional discussion and approaches can be found in survey articles by Barrett et al. \cite{barrett2023identifying} and Yao et al. \cite{yao2024survey}.\\

\noindent\textbf{Open challenges}: A key challenge is to ensure that robustness and security mechanisms are not intentionally or unintentionally removed in the process of finetuning an LLM \cite{qi2023fine}.
Another challenge lies in ensuring that the mechanisms work not just during evaluation but also during deployment (e.g., not subject to deceptive attacks \cite{hubinger2024sleeper}).
A broader challenge is to investigate robustness, security, and safety of systems that could be composed of multiple LLMs. For example, it has been shown that adversaries can misuse combinations of models by decomposing a malicious task into subtasks, leveraging aligned frontier models to solve hard but benign subtasks, and leveraging weaker non-aligned models to solve easy but malicious subtasks ~\cite{jones2024adversaries}. As such attacks do not require the aligned frontier models to generate malicious outputs and hence can go undetected, there is a need to extend red-teaming efforts beyond single models in isolation.

\subsection{Privacy, Unlearning, and Copyright Implications}
\noindent\textbf{Business problems}: How do we ensure that LLMs, diffusion models, and other generative AI models do not memorize training data instances (including personally identifiable information (PII)) and reproduce such data in their responses? How do we detect PII in LLM prompts / responses? How do prevent copyright infringement by LLMs?
How can we make an LLM / generative AI model forget specific parts, facts, or other aspects associated with the training data?\\

\noindent\textbf{Solution approaches}: Recent studies have shown that training data can be extracted from LLMs \cite{carlini2021extracting} and from diffusion models \cite{carlini2023extracting} (which could have copyright implications in case the model is perceived as a database from which the original images or other copyrighted data can be approximately retrieved). Several approaches for watermarking \cite{fairoze2023publicly,gu2022watermarking,kirchenbauer2023watermark} (or otherwise identifying / detecting \cite{mitchell2023detectgpt}) AI generated content have been proposed. Detecting PII in LLM prompts / responses can be done using off-the-shelf packages, but may require domain-specific modifications since what is considered as PII could vary based on the application. Unlearning in LLMs \cite{liu2024rethinking}, and more broadly, model disgorgement \cite{achille2023ai} (``the elimination of not just the improperly used data, but also the effects of improperly used data on any component of an ML model'') are likely to become important for copyright and privacy safeguards, ensuring responsible usage of intellectual property, compliance, and related requirements as well for reducing bias or toxicity and increasing fidelity.\\

\noindent\textbf{Open challenges}: A key challenge would be designing practical and scalable techniques. For example, how can we develop differentially private model training approaches (e.g., DPSGD~\cite{abadi2016deep}, PATE~\cite{papernot2018scalable})\footnote{Examples of differentially private model training include DPSGD \cite{abadi2016deep} and PATE\cite{papernot2018scalable}. While DPSGD operates by controlling the influence of training data during gradient descent, PATE transfers to a ``student'' model the knowledge of an ensemble of ``teacher'' models, with intuitive privacy provided by training teachers on disjoint data and strong privacy guaranteed by noisy aggregation of teachers' answers.} that are applicable for billions or trillions of parameters in generative AI models? How can we ensure privacy of end users when leveraging inputs from end users as part of retraining of LLMs (using, say, PATE-like approaches)? Considering the importance of high quality datasets for evaluating LLMs for truthfulness, bias, robustness, safety, and related dimensions, and the challenges with obtaining such datasets in highly sensitive domains such as healthcare, how do we develop practical and feasible approaches for differentially private synthetic data generation~\cite{aydore2021differentially,liu2021iterative,tao2021benchmarking}, potentially leveraging a combination of sensitive datasets (e.g., patient health records and clinical notes) and publicly available datasets along with the ability to generate data by querying powerful LLMs?

\subsection{Calibration and Confidence}
\noindent\textbf{Business problems}: How can we deploy LLMs in a human-AI hybrid setting to quantify the uncertainty (confidence score) associated with an LLM response and defer to humans when confidence is low? Specifically, how can we achieve this in high-stakes and latency-sensitive domains such as AI models used in healthcare settings?\\

\noindent\textbf{Solution approaches}: Learning to defer in human-AI settings is an active area of research \cite{keswani2021towards}, necessitating uncertainty quantification and confidence estimation for the underlying AI models. It also involves understanding the conditions under which humans can effectively complement AI models \cite{donahue2022human}. In the context of LLMs, recent approaches such as selective prediction, self-evaluation and calibration, semantic uncertainty, and self-evaluation-based selective prediction have been proposed \cite{chen2023adaptation} (see references there-in).\\

\noindent\textbf{Open challenges}:
A key challenge is to ensure that self-evaluation, calibration, selective prediction, and other confidence modeling approaches for LLMs are effective in out-of-distribution settings.
This is particularly important for adoption in high-stakes settings like healthcare.
Another challenge is ensuring robustness of confidence modeling approaches against adversarial prompts.

\subsection{Transparency and Causal Interventions}
\noindent\textbf{Business problems}: How do we explain the inner workings and responses of LLMs and other generative AI models, especially in scenarios requiring the development of end-user trust and meeting regulatory requirements? How can we modify factual associations linked to an LLM without retraining it?\\

\noindent\textbf{Solution approaches}: Explainability methods for LLMs have been well studied \cite{zhao2024explainability}, including techniques such as Chain-of-Thought Prompting \cite{wei2022chain} and variants. However, there is work on unfaithful explanations in chain-of-thought prompting \cite{turpin2023language}, with connections to language model alignment through externalized reasoning (getting models to do as much processing/reasoning through natural language as possible). Mechanistic interpretability \cite{rauker2023toward} is another active area of research, which has the potential to be further accelerated by the availability of small language models like phi-2. 
Causal tracing approaches have been proposed to locate and edit factual associations in LLMs. This involves first identifying neuron activations that are decisive in the model's factual predictions, and then modifying these neuron activations to update specific factual associations \cite{meng2022locating}.\\

\noindent\textbf{Open challenges}: Analogous to the use of simpler approximate models for explaining complex predictive ML models (e.g., LIME), can we employ simpler approximate models to explain LLMs and other generative AI models (e.g., using approaches such as model distillation) in a faithful manner? Additionally, can we develop more efficient and practical causal intervention approaches?

\section{Grounding for LLMs}\label{sec:grounding}
\noindent\textbf{Business problem}: How do we ensure that responses generated by an LLM are grounded in a user-specified knowledge base? 
Here, ``grounding'' means that every claim in the response can be attributed~\cite{ais_rashkin} to a document in the knowledge base.
We distinguish between the terms ``grounding'' and ``factuality''.
While ``grounding'' seeks attribution to a user-specific knowledge base, 
``factuality'' seeks attribution to commonly agreed world knowledge.

In the context of ``grounding'', the knowledge base may be a set of public and/or private documents, one or more Web domains, or the entire Web.
For instance, a healthcare company may want its chatbot to always produce responses that are grounded in a set of healthcare articles it consider authoritative.
In addition to grounding to the knowledge base, one may also want responses to contain citations into the relevant documents in the knowledge base.
This enables transparency and allows the end-user to corroborate all claims in the response.

\subsection{Solution Approaches}
In \S\ref{sec:truthfulness}, we laid out some key directions for detecting and preventing hallucinations in LLM responses.
As mentioned earlier, the requirement of grounding goes a step further from merely preventing hallucinations.
We seek responses that are fully aligned with a given knowledge base.
For instance, there may be a well-supported, non-hallucinated claim that disagrees with the provided knowledge base.
Such a claim would still be considered ungrounded.
There is a vast and growing literature on grounding for LLMs.
Below, we sketch out the key directions in this space.\\

\noindent\textbf{Retrieval Augmented Generation.}
Grounding failures often occur because not all information in the knowledge base is stored in the LLM's parametric memory.
One popular approach to circumventing this challenge is \emph{Retrieval Augmented Generation} (RAG)~\cite{rag_neurips20, rag_eacl21}, which leverages in-context learning to expose the model to relevant information from the knowledge base.
Specifically, given a prompt (user question), we retrieve relevant snippets (called \emph{context}) from the knowledge base, augment the prompt with this context, and then generate a response with the augmented prompt.
The success of a RAG system relies on the success of the retrieval step and the generation step. Consequently, RAG systems are evaluated based on dimensions such as {\em context relevance} (that is, whether the retrieved context is relevant to the given prompt), {\em answer faithfulness} (that is, whether the response generated by the LLM is properly grounded in the retrieved context), and {\em answer relevance} (that is, whether the response is relevant to the user question)~\cite{saad2024ares, es2024ragas}.

The retrieval step seeks to efficiently retrieval all relevant information for a given prompt.
This typically involves chunking and indexing the knowledge base into a vector database, and querying it based on the prompt.
The tremendous commercial interest in RAG systems has led to a proliferation of enterprise-grade vector databases (e.g., Pinecone \cite{pinecone}, FAISS \cite{douze2024faiss}) that enable retrieval from arbitrary knowledge bases.

To sharpen the retrieval step, several recent works have been exploring various aspects of it, including, the appropriate granularity of retrieval (such as a paragraph or a sentence) \cite{chen2023dense, long_context_rag}, strategies for decomposing complex prompts into one or more retrieval queries \cite{qi-etal-2023-art, adapative_rag, chan2024rqrag}, supervising retrieval based on quality of downstream generation systems \cite{wang2023learning}, and leveraging LLMs as retrieval indexes \cite{generate_then_read, transformer_search_index}.

The generation step seeks to ensure that the LLM's response remains grounded in the provided context.
This is not a given, especially when the context conflicts with the LLM's own parameteric knowledge \cite{rag_tug_of_war}.
To combat this, a common strategy is to fine-tune the LLM on \emph{(prompt, context, response)} triples~\cite{asai2024selfrag, luo2023search}.
In order to improve robustness, it is important to include a variety of contexts with varying levels of noise in the tuning set~\cite{luo2023search}.
To further incentivize LLMs to respond based on the provided context, recent work \cite{köksal2023hallucination, Kaushik2020Learning} proposes additional fine-tuning on counterfactual contexts and responses that contain claims that conflict with the model's parameter memory.
Besides fine-tuning, it is also possible to use reinforcement learning (RL) to reward grounded responses~\cite{gopher_cite}.
The reward model may be trained on human feedback on grounding, or may use automated models for performing checks (discussed below).

Another challenge for the generation step is comprehending contexts with temporal information.
For instance, consider a context specifying health records of a patient and the query: ``How does the patient's blood pressure from this week compare to last week?''
Producing a grounded response to this prompt requires knowing the current week, and identifying the blood pressures from the current week and the prior week.

Finally, it should be noted that no matter how effective the retrieval system is, there will always be instances of out-of-domain, adversarial, or nonsensical prompts where the retrieved context remains irrelevant. In such cases, it is crucial to train the model to generate an ``I don't know'' response by including demonstrations of such scenarios in the tuning set \cite{zhang-etal-2024-r, feng2024dont}.

Retrieval augmented generation is a vast area of research, and the above description provides only a brief overview.
We refer interested readers to surveys dedicated to RAG frameworks \cite{gao2024retrievalaugmented, zhao-etal-2023-retrieving, zhao2024retrievalaugmented}.\\

\noindent\textbf{Constrained Decoding.}
Another direction for improving groundedness is to use constrained decoding~\cite{yang-klein-2021-fudge, mudgal2024controlled, tu2024unlocking, NEURIPS2020_summarize_with_human_feedback}.
Here, the key idea is to modify the decoding strategy to optimize the groundedness of decoded responses.
A simple version of this is \emph{Best-of-N} sampling, wherein, we sample N different responses and select the ones with the largest grounding reward~\cite{gopher_cite, tu2024unlocking}.
Other works like FUDGE~\cite{yang-klein-2021-fudge} propose mechanisms for altering next word probabilities based on the likelihood of the current (partial) sequence completing into one that satisfies a certain attribute.
One can leverage these ideas to optimize for the grounding attribute~\cite{tu2024unlocking}.
Another direction is context-aware decoding~\cite{shi-etal-2024-trusting}, which upweights token probabilities to amplify the difference between generations with and without the provided context.
A common caveat for constrained decoding approaches is balancing groundedness with other desirable attributes like coherence, fluency, and helpfulness.\\

\noindent\textbf{Evaluation, Guardrails, and Revision.}
While the above directions make great strides towards improving grounding of LLM responses, they are not perfect.
For instance, multiple recent evaluations find that models struggle to generate grounded responses for prompts seeking time-varying information (e.g., ``Who won the latest soccer match between Liverpool and Manchester United?'')~\cite{fresh_qa}, and balancing grounding with other response attributes~\cite{gao-etal-2023-enabling, liu2023evaluating}.
In light of this, it is important to have inference time guardrails for verifying grounding.
There is extensive work on guardrails for LLM responses, to mitigate unsafe and harmful responses~\cite{inan2023llama, magooda2023framework} and copyright violations, and to protect against LLM vulnerabilities like prompt injection and jailbreaking \cite{rebedea-etal-2023-nemo, garak}.
Below, we specifically discuss guardrails for checking grounding of responses.

For responses generated by RAG frameworks, grounding verification is carried out by comparing the response to the context retrieved as part of the RAG retrieval step.
The most common way of doing this is to use a natural language inference (NLI) model to determine if the context entails the response~\cite{honovich2022true}.
Longer responses may be broken into individual sentences, and a separate NLI call may be made for each claim~\cite{gao2023rarr}.
This also allows identifying citations for each claim in the response.
The key advantage of this approach is that smaller T5-family~\cite{raffel_t5} models can be trained to perform NLI checks, making this approach attractive for inference-time grounding verification.
While NLI based checks are getting rapidly deployed as guardrails for grounding~\cite{gcp_cg, azure_cg}, they often struggle with performing grounding checks that involve reasoning.
Examples include validating claims making temporal statements, e.g., ``the patient's latest blood pressure is 130/80'', or claims involving negation, e.g., ``none of the reviews mention that the breakfast was bad'', or claims involving quantifiers, e.g., ``many guests appreciated the free breakfast''.

An alternative approach is to use an LLM to perform the grounding checks ~\cite{asai2024selfrag}.
This allows leveraging the LLM's superior world knowledge and reasoning abilities in making entailment judgements.
In general, LLM based approaches excel when the response is more abstract and does not quote directly from the context.
However, LLM based approaches are more computationally expensive making them less viable as inference time guardrails.

Finally, there is an emerging line of work on automatically and iteratively revising LLM responses in light of grounding feedback~\cite{madaan2023selfrefine, gao2023rarr, peng2023check, vernikos-etal-2024-small}.
Some approaches consider off-the-shelf LLMs to perform the revision tasks, while others train smaller, dedicated revision models~\cite{vernikos-etal-2024-small}.\\

\noindent\textbf{Corpus Tuning.}
An orthogonal approach to retrieval augmented generation is to pretrain the LLM on documents from the knowledge base to allow it to learn representations tailored to the knowledge base~\cite{gururangan-etal-2020-dont, xie2023data}.
This is particularly helpful when the knowledge base falls in a niche domain and/or involves novel terms not in the model's vocabulary; this is commonly the case for medical and healthcare domains.
Such domain-specific tuning is expected to benefit both closed book question-answering ~\cite{roberts-etal-2020-much} as well as RAG approaches~\cite{gururangan-etal-2020-dont}.

\subsection{Open Challenges}
Grounding for LLMs is a rapidly evolving area with several open challenges.
A key practical challenge for RAG frameworks is grappling with imperfect retrieval.
For instance, how should the model respond when the retrieval includes multiple opinions that contradict with each other, when the retrieval is missing crucial information sought by the prompt, or when the retrieval is completely irrelevant?
In some cases, even when the retrieval is missing information, the model may still have the necessary information in its parametric memory. How should models balance amongst answering from the context versus answering from parametric memory versus not answering at all (punting)?

A key challenge in tuning LLMs towards generating grounded responses is that the models may optimize for grounding at the expense of losing creativity and helpfulness.
For instance, they may quote verbatim from the provided context, which was recently observed for a number of commercial generative AI search engines~\cite{liu2023evaluating}.

Finally, a large open area is extending RAG frameworks to multimodal settings -- for instance, settings where the underlying knowledge base may consist of text, images, audio, and video, or the query may be a combination of text and audio.
This is an emerging area, and we refer interested readers to a recent survey by Zhao et al.~\cite{zhao-etal-2023-retrieving}.

\section{LLM Operations and Observability}\label{sec:observability}
\noindent\textbf{Business problems}: What processes and mechanisms are important for addressing grounding and evaluation related challenges in real-world LLM application settings in a holistic manner? How can we monitor LLMs and other generative AI applications deployed in production for metrics related to quality, safety, and other responsible AI dimensions? How can we anticipate and manage risks from frontier AI systems?

\subsection{Solution Approaches}
The emerging area of ``LLM operations'' deals with processes and tools for designing, developing, and deploying LLMs, as well as monitoring LLM applications once they are deployed in production. 
Frameworks such as the following have been proposed to address potential harms and challenges pertaining to grounding, robustness, and evaluation in real-world LLM applications \cite{raiAzure,gallegos2023bias,bender2021dangers}.
\begin{itemize}
    \item Identification \cite{hubinger2024sleeper,perez2022red,ganguli2022red}: Recognizing and prioritizing potential harms through iterative red-teaming, stress-testing, and thorough analysis of the AI system.
    \item Measurement \cite{guo2023evaluating,saad2024ares,magooda2023framework,zhu2023promptbench}: Establishing clear metrics, creating measurement test sets, and conducting iterative, systematic testing—both manual and automated—to quantify the frequency and severity of identified harms.
    \item Mitigation \cite{rebedea-etal-2023-nemo,inan2023llama,metzler2021rethinking,tamkin2023evaluating}: Implementing tools and strategies, such as prompt engineering and content filters, to reduce or eliminate potential harms. Repeated measurements need to be conducted to assess the effectiveness of the implemented mitigations. We could consider four layers of mitigation at model, safety system, application, and positioning levels.
    \item Operationalization \cite{gupta2023measuring,frontierAIRisk2023}: Defining and executing a deployment and operational readiness plan to ensure the responsible and ethical use of AI systems.
\end{itemize}

Depending on the domain requirements, an ``AI safety layer'' for detecting toxicity and other undesirable outputs in realtime can be included between the model and the application. Measuring shifts in the distribution of LLM prompts or responses could be helpful to identify potential degradation of the model quality over time, and further this information can be combined with any user feedback signals to determine regions where the model may be underperforming \cite{gupta2023measuring}.

Further, we need to differentiate undesirable outcomes or failures in LLM applications caused by adversarial attacks from failures due to the LLM's behavior in an unexpected manner in certain contexts. To address the latter class of ``unknown unknown'' failures, we should not only perform extensive testing and red teaming to preemptively identify and mitigate as many potential harms as possible but also incorporate processes and mechanisms to react quickly to any unanticipated harms during deployment. As an example, Microsoft introduced a new category of harms called ``Disparaging, Existential, and Argumentative'' harms as part of the responsible AI evaluation for conversational AI applications in response to the unexpected behavior of the Bing AI chatbot as reported by a New York Times journalist~\cite{roose2023conversation}.

More broadly, the risk profile associated with frontier AI systems is expected to expand in light of extensions of existing LLMs, e.g., multimodality, tool use, deeper reasoning and planning, larger and more capable memory, and increased interaction between AI systems \cite{frontierAIRisk2023,shevlane2023model}.
Of these, tool use is considered to create several new risks and vulnerabilities.

\subsection{Open Challenges}
A key challenge is to classify potential risks associated with tool use, AI agents, interaction between AI systems, etc., in terms of the level of attention and action needed now and at different points in the future.
This involves prioritizing investments to address such risks, especially in the following two areas: (1) Identifying failure modes and tendencies of LLM-based applications: We need to pinpoint how these applications can be led astray, and (2) Developing new safety and monitoring practices: This involves leveraging metrics like weight updates, activations, and robustness statistics, which are not currently available as part of LLM APIs. 

\section{Conclusion}
Given the increasing prevalence of AI technologies in our daily lives, it is crucial to integrate responsible AI methodologies into the development and deployment of Large Language Models (LLMs) and other Generative AI applications. We must understand the potential harms these models may introduce, and leverage  state-of-the-art techniques for enhancing overall quality, fairness, robustness, and explainability. Addressing the responsible AI related harms and challenges not only reduces legal, regulatory, and reputational risks, but also safeguards individuals, businesses, and society as a whole.
Moreover, there is a pressing need to establish ways to quantitatively assess the performance, quality, and safety of such models. Without comprehensive evaluations, establishing trust in LLM-based applications becomes exceedingly difficult. The goal of this tutorial is to establish a foundation for the development of safer and more reliable generative AI applications in the future.

\section*{Acknowledgments}
The authors would like to thank numerous researchers, practitioners, and industry leaders for insightful discussions which helped shape the business problems, solution approaches, and open challenges discussed in this article, and
Mark Johnson
and
Qinlan Shen
for thoughtful feedback.


{
\bibliographystyle{ACM-Reference-Format}
\balance
\bibliography{paper}
}

\end{document}